# Constructing a personalized AI assistant for shear wall layout using Stable Diffusion

Lufeng Wang[a,b], Jiepeng Liu[a,b], Guozhong Cheng[a,b,*], En Liu[c], Wei Chen[c]

[a] *Key Laboratory of New Technology for Construction of Cities in Mountain Area (Chongqing University), Ministry of Education, Chongqing 400045, China*

[b] *School of Civil Engineering, Chongqing University, Chongqing 400045, China*

[c] *China Construction Science & Technology Group Western Co., Ltd., Chongqing 401320, China*

**Abstract**

Shear wall structures are widely used in high-rise residential buildings, and the layout of shear walls requires many years of design experience and iterative trial and error. Currently, there are methods based on heuristic algorithms, but they generate results too slowly. Those based on Generative Adversarial Networks (GANs) or Graph Neural Networks (GNNs) can only generate single arrangements and require large amounts of training data. At present, Stable Diffusion is being widely used, and by using the Low-Rank Adaptation (LoRA) method to fine-tune large models with small amounts of data, good generative results can be achieved. Therefore, this paper proposes a personalized AI assistant for shear wall layout based on Stable Diffusion, which has been proven to produce good generative results through testing.

**Keywords:** Stable Diffusion; AI-generated Content; Personalized AI; Automated layout; Shear wall structure;

## 1. Introduction

The reinforced concrete shear wall systems have been widely used in high-rise buildings [1,2]. The shear wall layout plays a crucial role in earthquake resistance. However, the design of shear wall layout often needs years of design experience and constant trial and error. Therefore, some studies on automatic shear wall layout have been conducted. For instance, Zhang et al. [3] combined architectural layout and structural performance to generate shear wall layouts based on the modified evolutionary algorithm (GA). In another work, Gan et al. [4] applied parametric modelling and a novel GA optimization method to generate the structural topology and optimize member dimensions. Lou et al. [5] developed a shear wall layout optimization strategy for minimizing the structural weight with constraints on the story drift and period ratio based on the tabu search. Tafraout et al. [6] proposed an approach to optimize the wall layouts considering the performance of the slab and a set of general structural guidelines and seismic design rules. Lou et al. [7] combined a response surface methodology (RSM) with a discrete Particle Swarm Optimization (PSO) technique to optimize member sizes. The aforementioned studies primarily employ intelligent evolutionary algorithms. They usually need lots of time to make iterations for the optimal solution.

Over the past few years, with the development of artificial neural networks (ANN), several deep learning-based approaches have also been investigated. Pizarro et al. [8] used convolutional neural network (CNN) models by combining two independent floor plan prediction networks to generate the shear wall layouts. Liao et al. [9] employed generative adversarial networks (GAN) to generate shear wall layouts utilizing abstracted, semantically interpreted, classified, and parameterized data. For sake of improving the local design of shear wall layout, Zhao et al. [10] proposed an attention-enhanced GAN and generated more reasonable layouts in local zones, such as elevator shafts. In another work, Zhao et al. [11] utilized graph neural networks (GNNs) by representing a shear wall layout with a graph, which can ingeniously reflect the topological characteristics of shear wall layouts.

The aforementioned research can rapidly obtain design results by extracting the designer's experience. However, to achieve satisfactory generative outcomes, a considerable amount of paired data is often required for training neural networks. Moreover, the convergence of the network model typically necessitates continuous debugging and analysis by specialized researchers. Furthermore, the generated results usually yield only a single option, limiting the choices available to the designer.

Artificial Intelligence Generated Content (AIGC), encompassing natural language, music, and images, has experienced explosive progress recently, primarily attributed to the employment of large-scale models. In terms of natural language processing, the ChatGPT series [12–14] has achieved remarkable results. Moreover, the open-source Large Language Model Meta AI (LLaMA) [15,16] has the potential to aid individuals in developing their personalized reasoning assistants. In the field of image generation, Diffusion models [17] have found wide-ranging applications. DALL-E-2 [18] and Midjourney [19] generates images from natural language descriptions based on the user's prompts, which both achieved good results. Moreover, Stable Diffusion [20] not only has powerful image generation capabilities but also fosters a large open-source community [21], which helps users built their personalized AI.

In addition, numerous large model fine-tuning approaches such as Hypernetworks [22], DreamBooth [23], LoRA [24] , and others have helped the public obtain their own personalized AI design assistants. Especially, LoRA achieves impressive results by freezing the pre-trained model's weight parameters and adding a bypass operation that performs dimensionality reduction followed by an increase in dimensionality, simulating intrinsic rank, as shown in Fig. 1. It allows for low-cost fine-tuning of large models and yields satisfactory outcomes. However, Stable Diffusion and LoRA have not yet been applied in shear wall layout design.

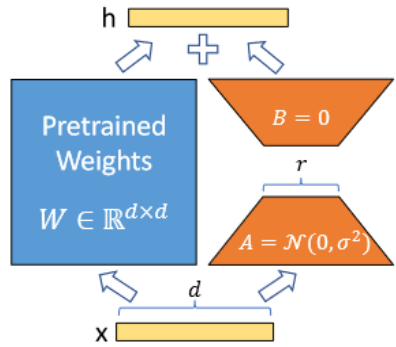

Fig. 1. LoRA (from [24])

To empower each structural designer with a personalized AI assistant for shear wall layout design, this paper presents a methodology framework based on Stable Diffusion. The main contributions of this work include: (1) proposing an automatic process for training the personalized AI assistant utilizing Stable Diffusion and LoRA, which necessitates only a small amount of preferred data; (2) providing a series of design parameters through experimental comparisons, which help generates reasonable layouts; and (3) presenting an automatic process for applying the personalized AI assistant using the Stable Diffusion Web-UI and Python, capable of generating diverse personalized layouts and being user-friendly in operation.

The remainder of this paper is organized as follows. Section 2 details the proposed methodology, including two parts: training of the LoRa network and application. Section 3 compares the design performance of the proposed method with existing design approaches. Finally, Section 4 concludes the paper and examines the research applicability.

## 2. Methodology

The structure of our proposed methodology framework for training and implementing a personalized AI assistant in shear wall layout design is depicted in Fig.2. This methodology principally comprises two phases: training the personalized AI assistant and applying the personalized AI assistant. The first phase necessitates collecting preferred data with the aid of the provided automatic pre-processor and training the LoRA network using this data. The application of the personalized AI assistant encompasses five steps: processing the architectural CAD; setting design parameters using the recommended values after trial; generating shear wall layouts with the previously trained LoRA; selecting the favored layout from the generated ones and making fine adjustments using PowerPoint (PPT); and utilizing the supplied post-processor for calculations in structural design software such as SAP2000 or PKPM.

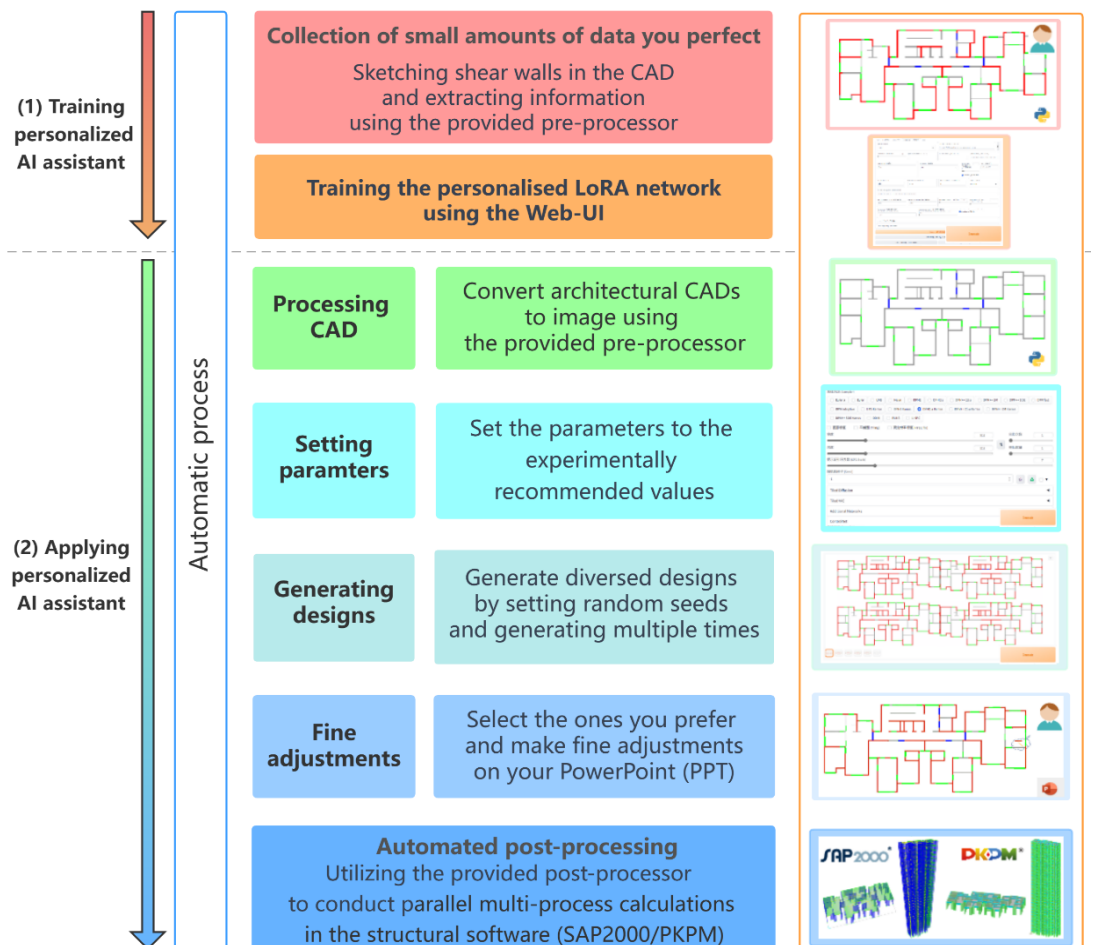

Fig. 2. Methodology framework for training and applying the personalized AI assistant in shear wall layout design

## 2.1 Training the personalized LoRA network

Fine-tuning the Stable Diffusion with the LoRA network typically doesn't require a substantial amount of data. However, pixelating approximately forty to fifty layout plans can also pose a challenge to ordinary users. Hence, this paper, based on [20], applies OpenCV [25] to pixelate the separately extracted geometries of the architectural floor plan and the shear wall floor plan, as illustrated in Fig. 3. Users can select their preferred drafts (approximately forty to fifty should suffice) and utilize the program provided in this study to automate the process of dataset creation.

Once the data is acquired, the LoRA can trained using the GUI for Stable Diffusion trainers [26]. Due to the powerful capabilities of LoRA, numerous parameters need to be adjusted in the GUI. The size of the training images can be adjusted based on the dimensions of the input images. It should be noted that to reduce the difficulty of training, users can train drawings of the same subcategory together, such as high-rise shear wall structures in the seventh-degree area. Therefore, all labels can be the same, such as "Seventh Degree High-Rise Building Shear Wall Structure" or "Shear Wall Layout", and so on. Through experimentation, it has been found that satisfactory results can be achieved with twenty epochs of training, each consisting of 100 steps.

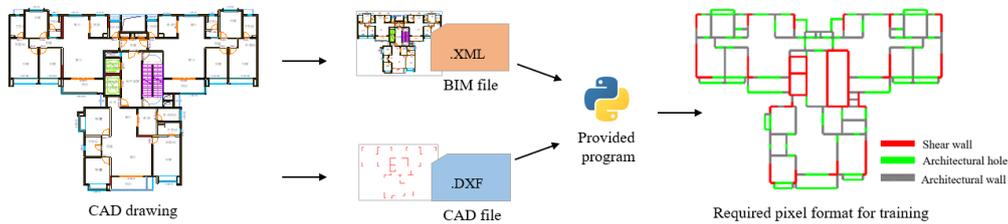

Fig. 3. Process of obtaining training images

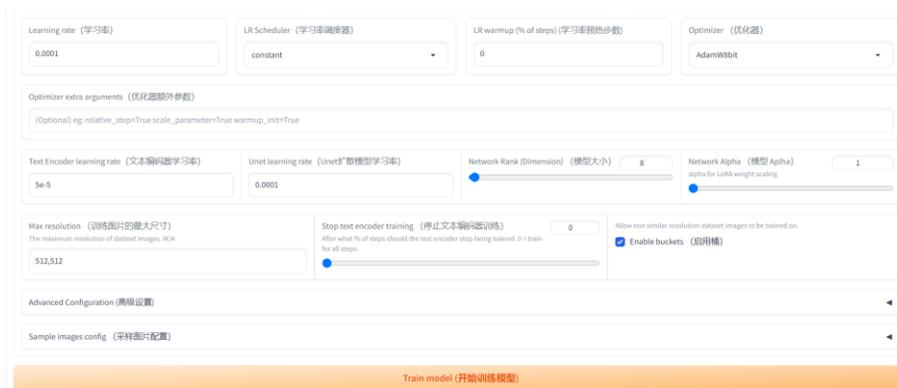

Fig. 4. GUI for Stable Diffusion trainers (reproduce from [26])

## 2.2. Application
### 2.2.1 Processing architectural CADs

Similar to the process outlined in Section 2.1 for obtaining training images, this step only requires the architectural CAD drawings to be processed. Users can employ the provided program to convert these CAD drawings into the necessary pixel images for the generation of shear walls, as shown in Fig. 5.

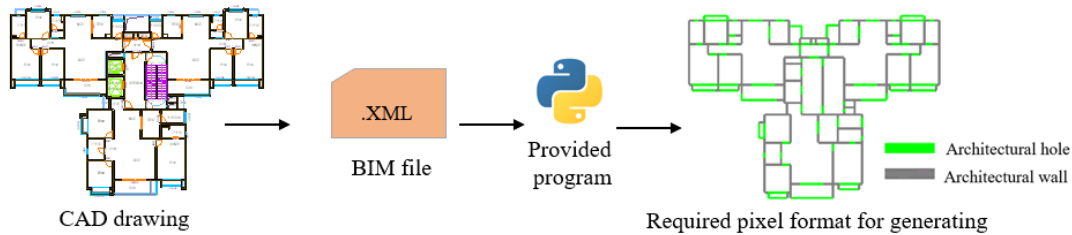

Fig. 5. Flowchart of getting required pixel format

### 2.2.2 Design parameters and generating designs

By utilizing the pre-trained personalized LoRA model, users can swiftly generate shear wall layouts through the Stable Diffusion Web-UI [27] by importing the generated pixel images, as shown in Fig. 6. It's important to note that underfitting or overfitting may occur during the training process, thus saving the LoRA model at different epochs can assist users in selecting the most suitable fine-tuned model. Through extensive experimentation, the authors have found that Sampler and ControlNet are crucial for the shear wall layout.

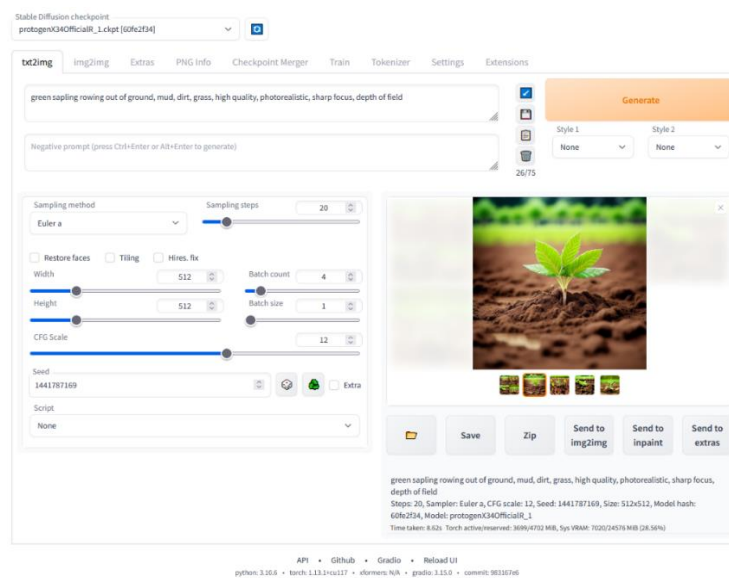

Fig. 6. Stable Diffusion Web-UI (from [27])

### 2.2.3. Fine adjustments and automated post-processing

Upon receiving generated images from the AI assistant, designers can promptly make fine adjustments to achieve their desired layout plans. This study proposes two methods for designers to make these adjustments (see Fig. 7.). Method 1 involves using PowerPoint (PPT) to place red blocks of the same width as the walls in the desired locations, then combining them and saving as an image. By utilizing the provided conversion program, the modeling in structural calculation software can be completed. Alternatively, Method 2 allows for direct modeling in structural calculation software using the provided conversion program, with adjustments made within the software itself. The calculation software commonly used in China, such as PKPM, and internationally, like SAP2000, can facilitate this process.

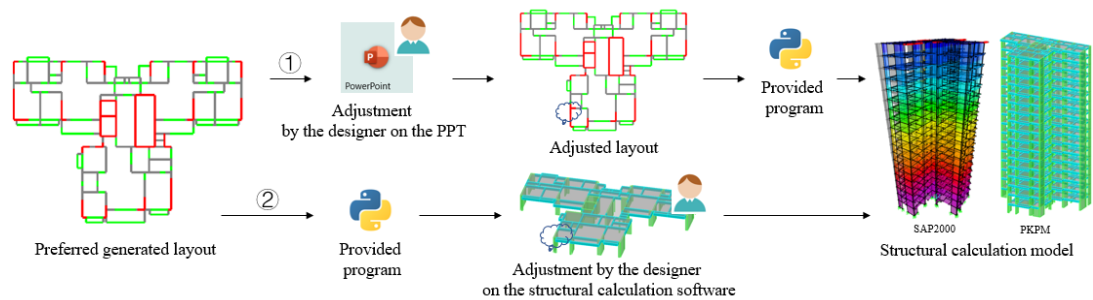

Fig. 7. Flowchart of obtaining required pixel format

### 2.2.4. Evaluation metrics

Structural design is a complex and rigorous task. In shear wall structures, the plan layout of shear walls significantly affects the seismic performance of the structure [28]. To validate the proposed method and relevant parameters in this study, a series of evaluation metrics are introduced.

First, each country has specific seismic design codes for earthquake resistance . In this study, taking the related Chinese code requirements as an example, five global seismic structural indicators are used as important evaluation metrics, as shown in Table 1. $\delta_{drift}$ represents the inter-story drift angle, which restricts the horizontal displacement of structures under normal usage conditions to ensure the required stiffness of high-rise structures, preventing excessive displacement that may affect the structure's load-bearing capacity, stability, and usage requirements. $r_{torsion}$ refers to the torsional ratio, which serves as an important basis for determining the presence and degree of torsional irregularity in structures. $r_{period}$ stands for the period ratio, which also controls the relative relationship between lateral stiffness and torsional stiffness, making the plan layout of lateral force-resisting elements more effective and rational, and preventing the structure from experiencing excessive torsional effects (relative to lateral displacement).

Table 1
Three global seismic structural indicators

| Item | Name | Description | Code limit |
| --- | --- | --- | --- |
| 1 | $1/\delta_{drift}$ | Inter-story drift angle | $\leq 1/1000$ |
| 2 | $r_{torsion}$ | Torsional ratio | $\leq 1.4$ |
| 3 | $r_{period}$ | Period ratio | $\leq 0.9$ |

In addition to the global seismic structural indicators, the planar geometric shape of the shear wall segments also greatly affects the seismic capacity of the structure. The presence of irregular columns complicates the force distribution and makes it difficult to predict, leading to local stress concentrations, which in turn can cause bending, twisting, and buckling instability, affecting the overall structural stability. Such irregularities should be avoided in typical shear wall structures. Similarly, short-limb shear walls (Fig. 8) have poor seismic performance and are prone to cracking under horizontal forces, so they should be avoided as much as possible. The aforementioned two indicators are represented by $N_{column}$ and $N_{short}$, as shown in Table 2. Besides, rectangular columns also should be avoided, and they are also accounted for in $N_{column}$. Furthermore, material consumption is an essential aspect of design and can be approximated by the total wall length ($L_{wall}$) of the shear walls.

Apart from the qualitative metrics mentioned above, the shear wall layout must be reasonable, which requires a level of expertise achievable only through years of design experience. Therefore, designers with multiple years of experience will provide a comprehensive score, ranging from 0 to 10, as listed in Table 2.

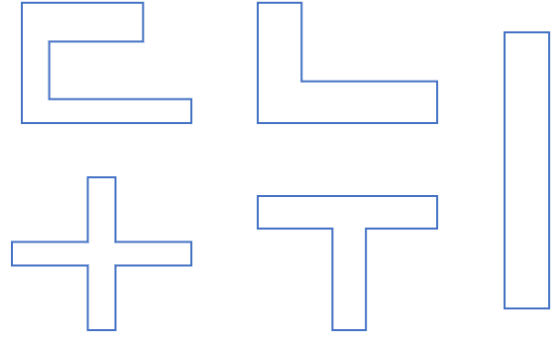

Fig. 8. Different plane geometries of short-limb shear walls

Table 2
Structural indicators

| Item | Name | Description |
|---|---|---|
| 4 | $N_{column}$ | Number of irregular and rectangular columns |
| 5 | $N_{short}$ | Number of short-limb shear walls |
| 6 | $L_{wall}$ | Total length of shear walls |
| 7 | $S_{layout}$ | Rationality of the layout, ranging from 0 to 10 |

## 3. Discussions of design performance

To compare with the studies [10,11], the dataset used for LoRA network training is part L1_7 of the open-source dataset [29]. To validate the proposed method without being overly influenced by specialized fine-tuning, a second-year graduate student in the field of architectural structure serves as the user for the selection and fine-tuning process. The input architectural floor plan is shown in Fig. 9. The floor plans generated by GAN and GNN are shown in Fig. 10. The floor plans generated by AIassist are shown in Fig. 11.

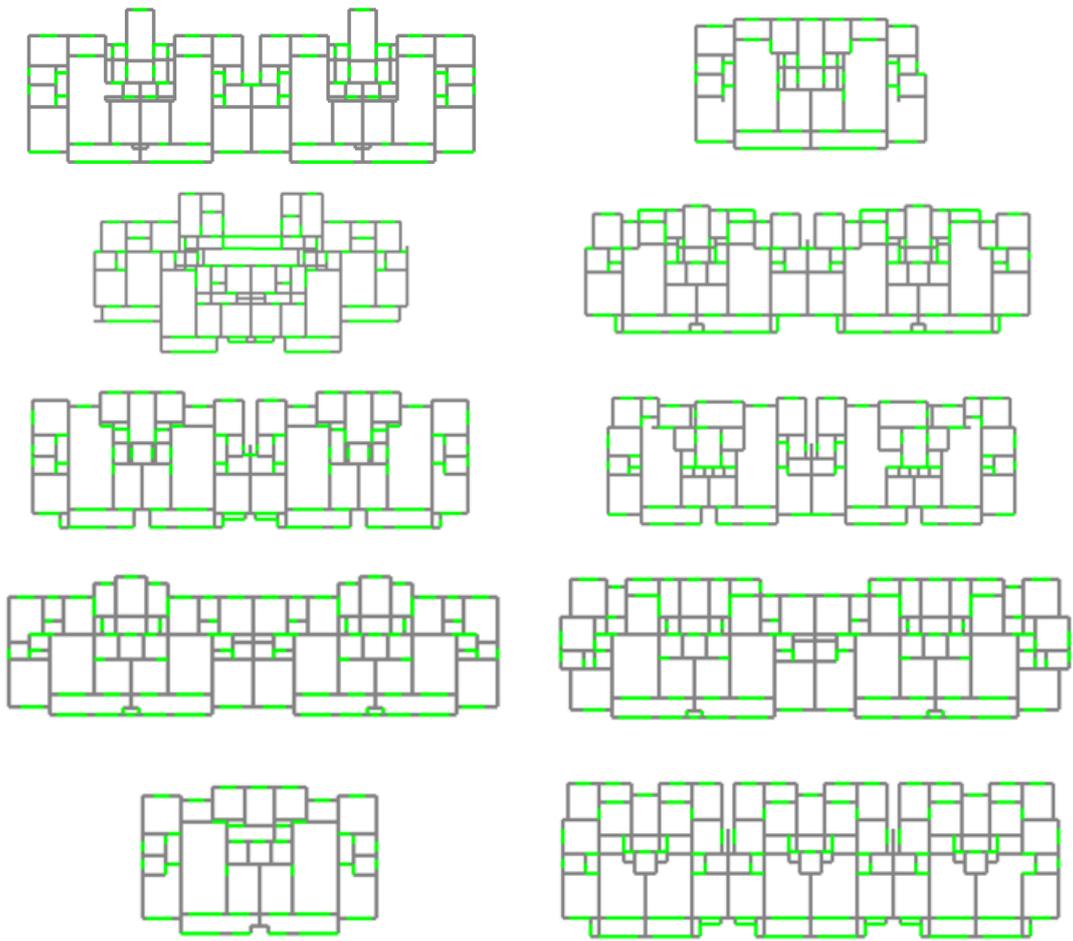

Fig. 9. Cases for comparison (2× 5)

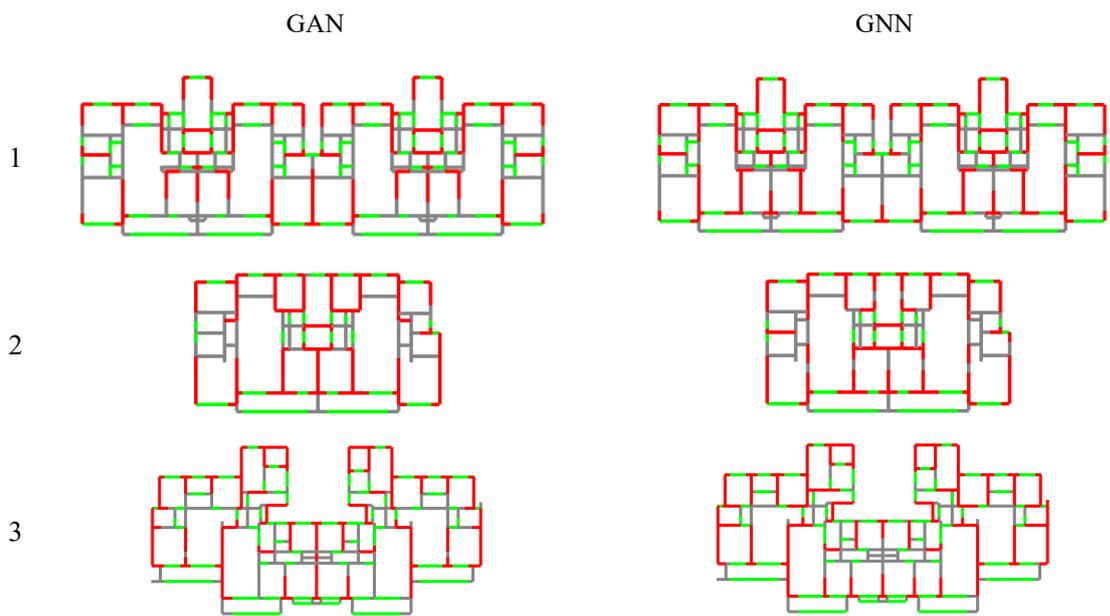

GAN  GNN

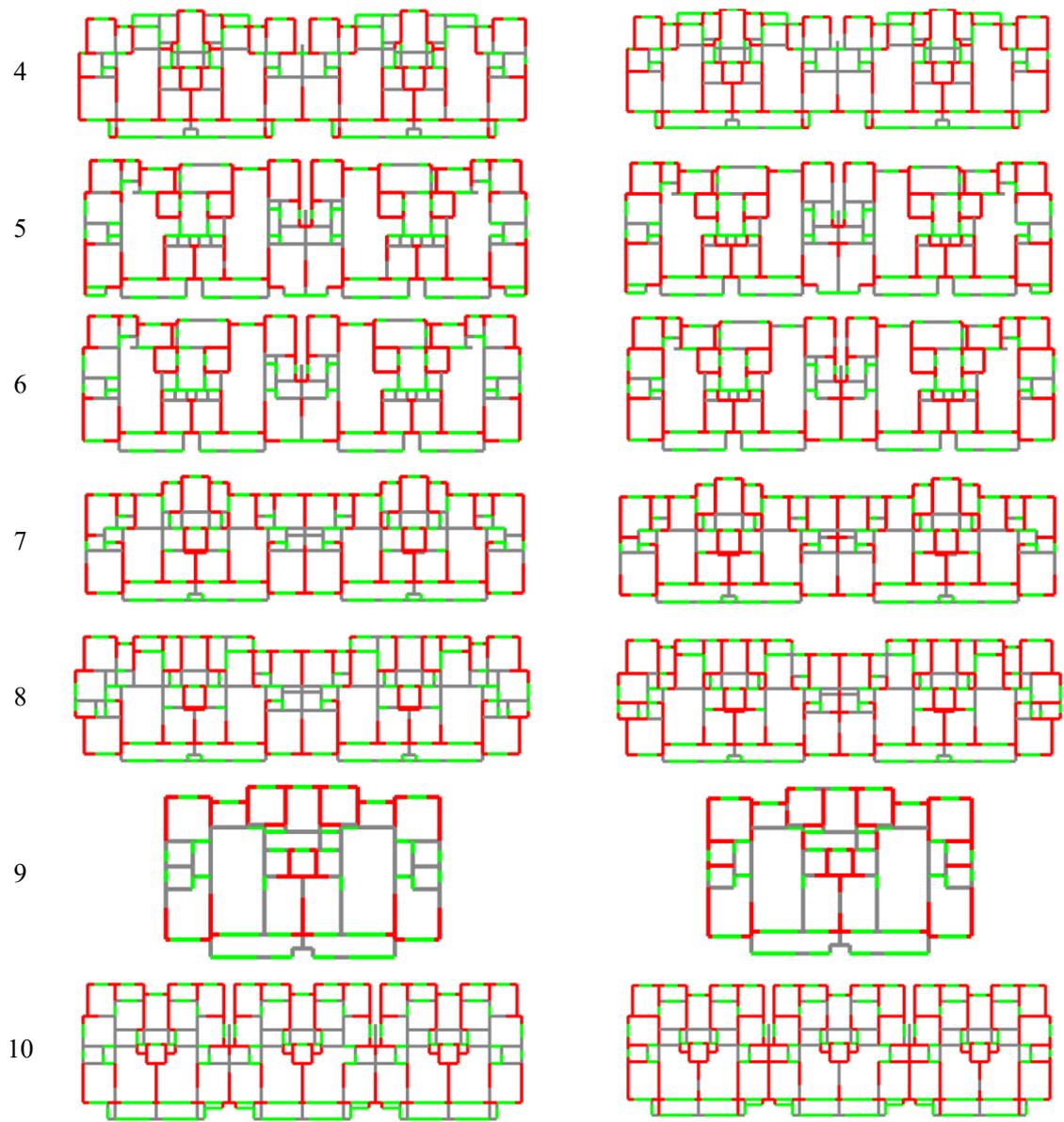

Fig. 10. Layouts generated by GAN [10] and GNN [11]

|   | Layout 1 | Layout 2 | Preferred layout | Adjusted layout |
|---|---|---|---|---|
| 1 | 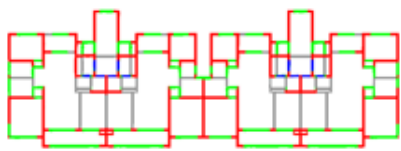 | 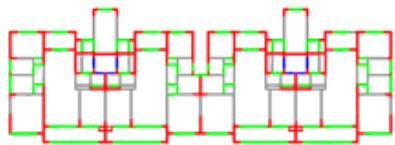 | 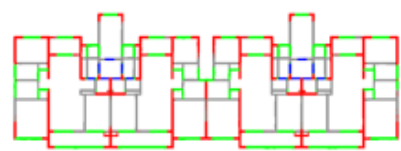 | 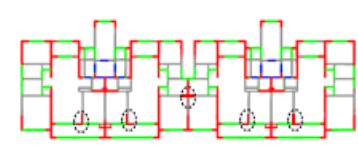 |
| 2 | 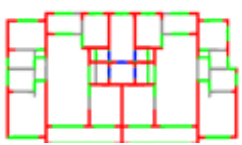 | 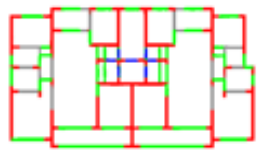 | 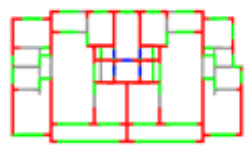 | 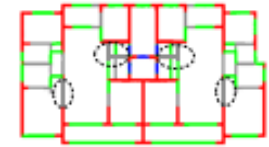 |
| 3 | 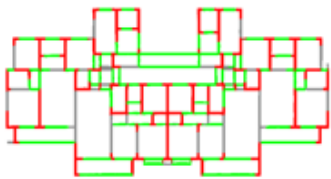 | 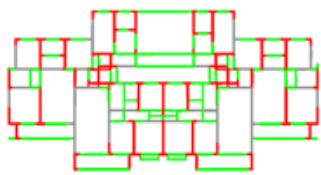 | 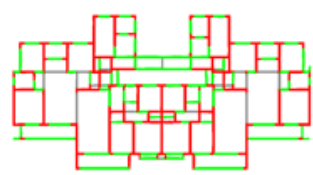 | 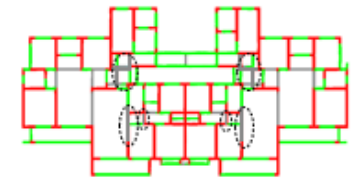 |
| 4 | 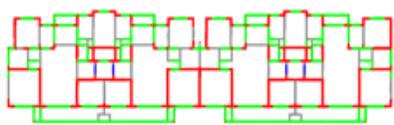 | 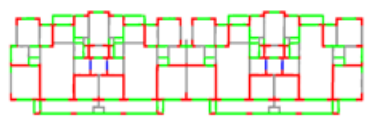 | 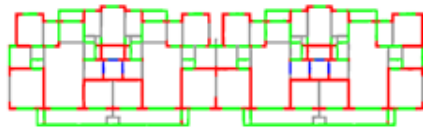 | 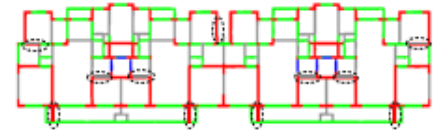 |
| 5 | 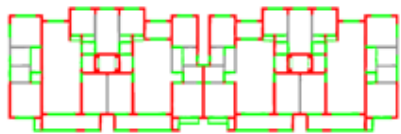 | 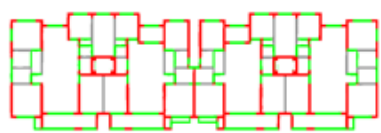 | 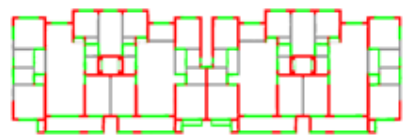 | 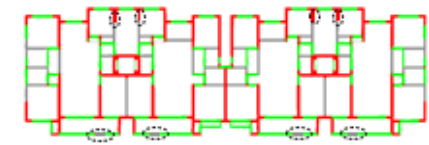 |

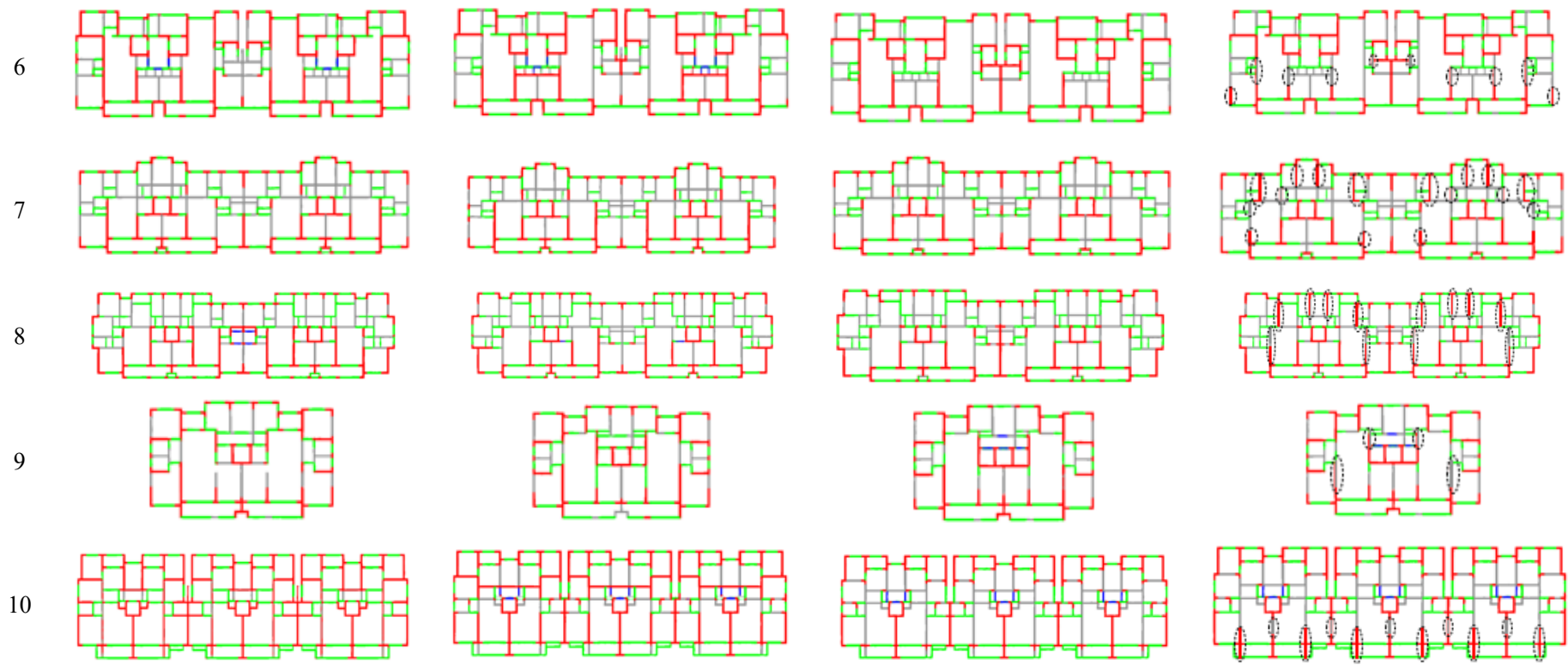

Fig. 11. Layouts generated by the AI assistant and adjusted by the designer

From the scores of critics (Fig. 12), it can be seen that the total length of shear walls in the proposed method is shorter, and both arrangements generally have more columns and short-limb shear walls. The shear walls become more reasonable after fine-tuning.

| Case | Critic | Score | | | |
|---|---|---|---|---|---|
| | | GAN | GNN | Preferred (AiAssist) | Adjusted |
| 1 | 1 | 4.5 | 5.0 | 6.0 | 6.0 |
| | 2 | 5.0 | 9.0 | 7.0 | 8.0 |
| | 3 | 2.0 | 5.0 | 5.0 | 5.0 |
| 2 | 1 | 5.0 | 5.0 | 6.2 | 6.5 |
| | 2 | 6.0 | 7.0 | 8.0 | 8.0 |
| | 3 | 4.0 | 4.0 | 6.0 | 6.0 |
| 3 | 1 | 5.0 | 4.5 | 4.0 | 4.0 |
| | 2 | 8.0 | 9.0 | 6.0 | 6.0 |
| | 3 | 8.0 | 8.0 | 6.0 | 6.0 |
| 4 | 1 | 5.0 | 5.0 | 5.5 | 5.5 |
| | 2 | 8.0 | 9.0 | 6.0 | 6.0 |
| | 3 | 9.0 | 9.0 | 6.0 | 6.0 |
| 5 | 1 | 6.5 | 6.5 | 5.8 | 5.8 |
| | 2 | 9.0 | 8.0 | 6.0 | 6.0 |
| | 3 | 9.0 | 9.0 | 6.0 | 6.0 |
| 6 | 1 | 6.0 | 6.0 | 5.5 | 6.2 |
| | 2 | 8.0 | 9.0 | 6.0 | 6.0 |
| | 3 | 7.0 | 7.0 | 5.0 | 5.0 |
| 7 | 1 | 5.5 | 5.5 | 5.8 | 5.8 |
| | 2 | 8.0 | 9.0 | 6.0 | 7.0 |
| | 3 | 8.0 | 8.0 | 5.0 | 7.0 |
| 8 | 1 | 6.5 | 6.5 | 5.5 | 6.0 |
| | 2 | 9.0 | 8.0 | 6.0 | 6.0 |
| | 3 | 9.0 | 8.0 | 6.0 | 7.0 |
|    | 1   | 6.5  | 6.5  | 6.5  | 6.5  |
| 9  | 2   | 8.0  | 9.0  | 6.0  | 6.0  |
|    | 3   | 7.0  | 8.0  | 7.0  | 7.0  |
|    | 1   | 5.5  | 5.5  | 6.0  | 5.8  |
| 10 | 2   | 8.0  | 8.0  | 6.0  | 6.0  |
|    | 3   | 7.0  | 7.0  | 6.0  | 7.0  |
| AVG. |   | 6.77 | 7.13 | 5.92 | 6.17 |

Fig. 12. Score of critics

Fig. 13 displays the values of various evaluation indicators, where orange-red signifies values exceeding the limit and green represents the minimum values among the corresponding indicators. As the Fig. 14 shows, the majority of the floor plans meet the requirements set by the standard limits. While the layouts generated by AI feature shorter shear walls, they also tend to include more columns and short-limb walls.

| Case | Generator | 1    | 2   | 3   | 4   | 5   | 6   |
|------|-----------|------|-----|-----|-----|-----|-----|
|      |           | 1000 | 1.4 | 0.9 | nan | nan | nan |
| 1    | GAN       | 3494 | 1.3 | 1.0 | 29  | 20  | 155 |
|      | GNN       | 2909 | 1.3 | 0.9 | 29  | 22  | 180 |
|      | AiAssist  | 2710 | 1.3 | 0.9 | 31  | 22  | 144 |
|      | Adjusted  | 3359 | 1.3 | 0.9 | 33  | 33  | 164 |
| 2    | GAN       | 3034 | 1.5 | 0.6 | 15  | 15  | 86  |
|      | GNN       | 2578 | 1.4 | 0.6 | 13  | 14  | 84  |
|      | AiAssist  | 4305 | 1,3 | 0.7 | 18  | 14  | 104 |
|      | Adjusted  | 3221 | 1.2 | 0.7 | 21  | 15  | 91  |
| 3    | GAN       | 3833 | 1.2 | 0.6 | 16  | 28  | 162 |
|      | GNN       | 3954 | 1.2 | 0.7 | 16  | 19  | 177 |
|      | AiAssist  | 5096 | 1.2 | 0.8 | 42  | 33  | 195 |
|      | Adjusted  | 5086 | 1.2 | 0.8 | 42  | 33  | 195 |
| 4    | GAN       | 4707 | 1.3 | 0.7 | 30  | 24  | 192 |

|   | Method | | | | | | |
|---|---|---|---|---|---|---|---|
|   | GNN | 4830 | 1.2 | 0.6 | 34 | 25 | 197 |
|   | AiAssist | 2997 | 1.2 | 0.8 | 46 | 31 | 149 |
|   | Adjusted | 3186 | 1.2 | 0.8 | 47 | 28 | 146 |
|   | GAN | 3868 | 1.4 | 0.9 | 32 | 30 | 165 |
| 5 | GNN | 3937 | 1.2 | 0.8 | 32 | 25 | 168 |
|   | AiAssist | 4177 | 1.3 | 0.8 | 27 | 20 | 161 |
|   | Adjusted | 3615 | 1.3 | 0.8 | 27 | 20 | 162 |
|   | GAN | 3908 | 1.3 | 0.8 | 21 | 30 | 174 |
| 6 | GNN | 4569 | 1.3 | 0.8 | 20 | 26 | 199 |
|   | AiAssist | 4088 | 1.2 | 0.8 | 33 | 21 | 161 |
|   | Adjusted | 3157 | 1.3 | 0.9 | 32 | 20 | 146 |
|   | GAN | 4248 | 1.2 | 0.8 | 28 | 25 | 173 |
| 7 | GNN | 4226 | 1.2 | 0.7 | 33 | 19 | 191 |
|   | AiAssist | 3012 | 1.3 | 0.8 | 51 | 30 | 121 |
|   | Adjusted | 3919 | 1.3 | 0.9 | 43 | 29 | 146 |
|   | GAN | 4398 | 1.2 | 0.8 | 24 | 24 | 183 |
| 8 | GNN | 4765 | 1.2 | 0.8 | 27 | 30 | 221 |
|   | AiAssist | 3254 | 1.2 | 0.8 | 46 | 47 | 140 |
|   | Adjusted | 4152 | 1.3 | 0.8 | 46 | 35 | 169 |
|   | GAN | 1811 | 1.4 | 0.6 | 9 | 7 | 68 |
| 9 | GNN | 2538 | 1.3 | 0.6 | 9 | 10 | 82 |
|   | AiAssist | 4327 | 1.3 | 0.7 | 27 | 11 | 87 |
|   | Adjusted | 3277 | 1.3 | 0.7 | 27 | 9 | 78 |
|   | GAN | 4495 | 1.3 | 0.9 | 21 | 46 | 218 |
| 10 | GNN | 5134 | 1.3 | 0.8 | 21 | 49 | 278 |
|   | AiAssist | 4131 | 1.2 | 0.8 | 70 | 37 | 197 |
|   | Adjusted | 4117 | 1.2 | 0.8 | 64 | 44 | 209 |

Fig. 13. Score of metrics

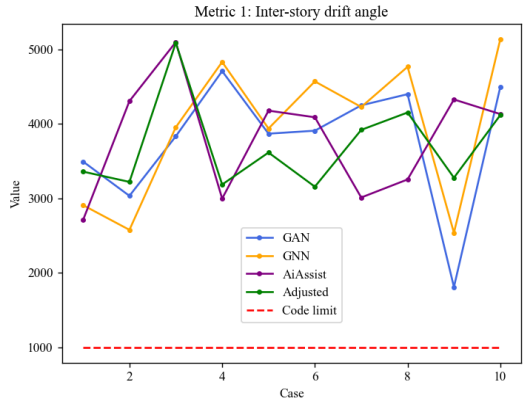

(a) Inter-story drift angle

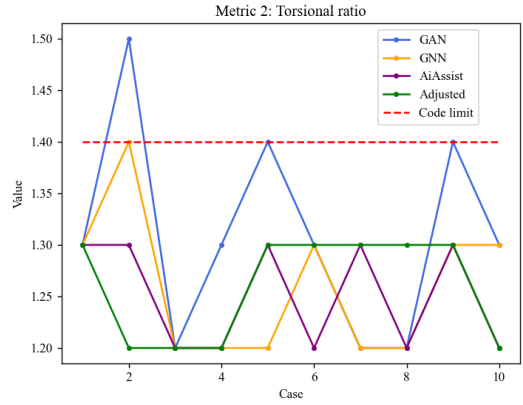

(b) Torsional ratio

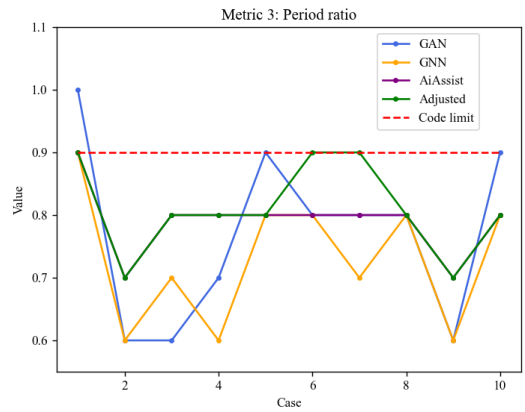

(c) Period ratio

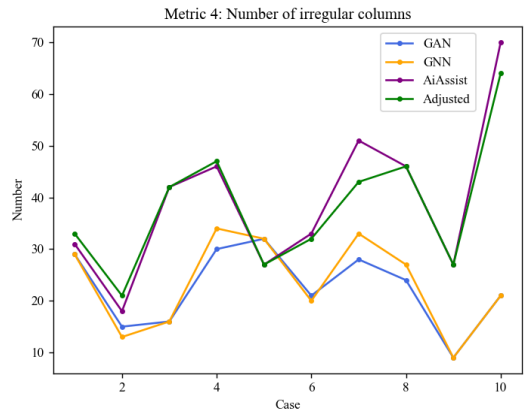

(d) Number of irregular and rectangular columns

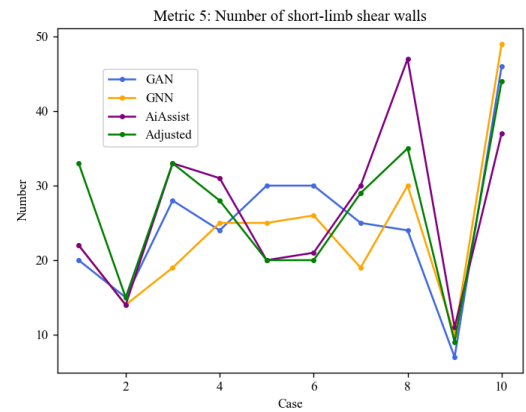

(e) Number of short-limb shear walls

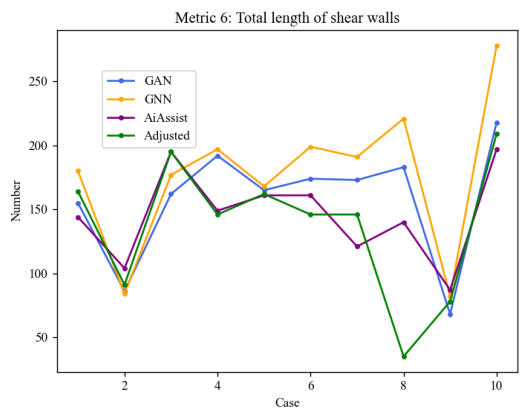

(f) Total length of shear walls

Fig. 14. Comparison of each metric

## 4. Conclusion and future work

The experiments conducted in this study have demonstrated that the proposed method can effectively assist designers in their work. Additionally, users can follow the procedures and open-source software provided in this paper to train and improve their own AI models.

**Reference**

[1] Z. Wang, W. Pan, Z. Zhang, High-rise modular buildings with innovative precast concrete shear walls as a lateral force resisting system, Structures. 26 (2020) 39–53. https://doi.org/10.1016/j.istruc.2020.04.006.

[2] H. Hu, J. Liu, G. Cheng, Y. Ding, Y.F. Chen, Seismic behavior of hybrid coupled shear wall with replaceable U-shape steel coupling beam using terrestrial laser scanning, Advances in Structural Engineering. 25 (2022) 1167–1177. https://doi.org/10.1177/13694332211065505.

[3] Y. Zhang, C. Mueller, Shear wall layout optimization for conceptual design of tall buildings, Engineering Structures. 140 (2017) 225–240. https://doi.org/10.1016/j.engstruct.2017.02.059.

[4] V.J.L. Gan, C.L. Wong, K.T. Tse, J.C.P. Cheng, I.M.C. Lo, C.M. Chan, Parametric modelling and evolutionary optimization for cost-optimal and low-carbon design of high-rise reinforced concrete buildings, Advanced Engineering Informatics. 42 (2019) 100962. https://doi.org/10.1016/j.aei.2019.100962.

[5] H. Lou, B. Gao, F. Jin, Y. Wan, Y. Wang, Shear wall layout optimization strategy for high-rise buildings based on conceptual design and data-driven tabu search, Computers & Structures. 250 (2021) 106546. https://doi.org/10.1016/j.compstruc.2021.106546.

[6] S. Tafraout, N. Bourahla, Y. Bourahla, A. Mebarki, Automatic structural design of RC wall-slab buildings using a genetic algorithm with application in BIM environment, Automation in Construction. 106 (2019) 102901. https://doi.org/10.1016/j.autcon.2019.102901.

[7] H. Lou, Z. Xiao, Y. Wan, G. Quan, F. Jin, B. Gao, H. Lu, Size optimization design of members for shear wall high-rise buildings, Journal of Building Engineering. 61 (2022) 105292. https://doi.org/10.1016/j.jobe.2022.105292.

[8] P.N. Pizarro, L.M. Massone, F.R. Rojas, R.O. Ruiz, Use of convolutional networks in the conceptual structural design of shear wall buildings layout, Engineering

Structures. 239 (2021) 112311. https://doi.org/10.1016/j.engstruct.2021.112311.

[9] W. Liao, X. Lu, Y. Huang, Z. Zheng, Y. Lin, Automated structural design of shear wall residential buildings using generative adversarial networks, Automation in Construction. 132 (2021) 103931. https://doi.org/10.1016/j.autcon.2021.103931.

[10] P. Zhao, W. Liao, Y. Huang, X. Lu, Intelligent design of shear wall layout based on attention-enhanced generative adversarial network, Engineering Structures. 274 (2023) 115170. https://doi.org/10.1016/j.engstruct.2022.115170.

[11] P. Zhao, W. Liao, Y. Huang, X. Lu, Intelligent design of shear wall layout based on graph neural networks, Advanced Engineering Informatics. 55 (2023) 101886. https://doi.org/10.1016/j.aei.2023.101886.

[12] A. Radford, K. Narasimhan, T. Salimans, I. Sutskever, Improving Language Understanding by Generative Pre-Training, (n.d.).

[13] A. Radford, J. Wu, R. Child, D. Luan, D. Amodei, I. Sutskever, Language Models are Unsupervised Multitask Learners, (n.d.).

[14] T.B. Brown, B. Mann, N. Ryder, M. Subbiah, J. Kaplan, P. Dhariwal, A. Neelakantan, P. Shyam, G. Sastry, A. Askell, S. Agarwal, A. Herbert-Voss, G. Krueger, T. Henighan, R. Child, A. Ramesh, D.M. Ziegler, J. Wu, C. Winter, C. Hesse, M. Chen, E. Sigler, M. Litwin, S. Gray, B. Chess, J. Clark, C. Berner, S. McCandlish, A. Radford, I. Sutskever, D. Amodei, Language Models are Few-Shot Learners, (2020). http://arxiv.org/abs/2005.14165 (accessed May 11, 2023).

[15] H. Touvron, T. Lavril, G. Izacard, X. Martinet, M.-A. Lachaux, T. Lacroix, B. Rozière, N. Goyal, E. Hambro, F. Azhar, A. Rodriguez, A. Joulin, E. Grave, G. Lample, LLaMA: Open and Efficient Foundation Language Models, (2023). http://arxiv.org/abs/2302.13971 (accessed May 11, 2023).

[16] LLaMA, (2023). https://github.com/facebookresearch/llama (accessed May 11, 2023).

[17] J. Ho, A. Jain, P. Abbeel, Denoising Diffusion Probabilistic Models, (2020). http://arxiv.org/abs/2006.11239 (accessed May 17, 2023).

[18] A. Ramesh, M. Pavlov, G. Goh, S. Gray, C. Voss, A. Radford, M. Chen, I. Sutskever, Zero-Shot Text-to-Image Generation, in: Proceedings of the 38th International Conference on Machine Learning, PMLR, 2021: pp. 8821–8831. https://proceedings.mlr.press/v139/ramesh21a.html (accessed May 11, 2023).

[19] Midjourney, Midjourney. (n.d.). https://www.midjourney.com/home/?callbackUrl=%2Fapp%2F (accessed May 17,

2023).

[20] R. Rombach, A. Blattmann, D. Lorenz, P. Esser, B. Ommer, High-Resolution Image Synthesis with Latent Diffusion Models, in: 2022 IEEE/CVF Conference on Computer Vision and Pattern Recognition (CVPR), 2022: pp. 10674–10685. https://doi.org/10.1109/CVPR52688.2022.01042.

[21] CompVis/stable-diffusion: A latent text-to-image diffusion model, (n.d.). https://github.com/CompVis/stable-diffusion (accessed May 11, 2023).

[22] D. Ha, A. Dai, Q.V. Le, HyperNetworks, (2016). http://arxiv.org/abs/1609.09106 (accessed May 11, 2023).

[23] N. Ruiz, Y. Li, V. Jampani, Y. Pritch, M. Rubinstein, K. Aberman, DreamBooth: Fine Tuning Text-to-Image Diffusion Models for Subject-Driven Generation, (2023). http://arxiv.org/abs/2208.12242 (accessed May 11, 2023).

[24] E.J. Hu, Y. Shen, P. Wallis, Z. Allen-Zhu, Y. Li, S. Wang, L. Wang, W. Chen, LoRA: Low-Rank Adaptation of Large Language Models, (2021). http://arxiv.org/abs/2106.09685 (accessed May 11, 2023).

[25] opencv-python: Wrapper package for OpenCV python bindings., (n.d.). https://github.com/opencv/opencv-python (accessed April 10, 2023).

[26] bmaltais, Kohya's GUI, (2023). https://github.com/bmaltais/kohya_ss (accessed May 10, 2023).

[27] AUTOMATIC1111/stable-diffusion-webui: Stable Diffusion web UI, (n.d.). https://github.com/AUTOMATIC1111/stable-diffusion-webui/tree/master (accessed May 10, 2023).

[28] X. Zhou, L. Wang, J. Liu, G. Cheng, D. Chen, P. Yu, Automated structural design of shear wall structures based on modified genetic algorithm and prior knowledge, Automation in Construction. (2022) 12. https://doi.org/10.1016/j.autcon.2022.104318.

[29] Wenjie Liao, StructGAN_v1, (2023). https://github.com/wenjie-liao/StructGAN_v1 (accessed May 15, 2023).